# Translating Classical Poetry into Modern Prose

**Chalamalasetti Kranti**
Computational Linguistics,
University of Potsdam, Germany
kranti.chalamalasetti@uni-potsdam.de

**Sowmya Vajjala**
National Research Council,
Ottawa, Canada
sowmya.vajjala@nrc-cnrc.gc.ca

## Abstract

We built a dataset for the task of poem-to-prose translation from 13th-17th Century Telugu classical poetry to contemporary Telugu and English prose, which we call పద్యం2గద్యం (PADYAM2GADYAM [1]). The dataset consists of 600 poems and their human-verified Telugu and English prose translations. We evaluated 2 machine translation systems and 5 contemporary Large Language Models (LLMs) on their ability to do zero-shot poem-to-prose translation into Telugu and English using this dataset. Our results indicate that while the general purpose LLMs are better than the machine translation systems, there are systematic issues with the generation and evaluation of prose translation in both languages across systems.

## 1 Introduction

*The poet moves from life to language, the translator moves from language to life; both, like the immigrant, try to identify the invisible, what's between the lines, the mysterious implications.*
– Anne Michaels, in Fugitive Pieces

The challenge of translating literature lies in identifying what is present not only in the words, but also between the lines. This is more pronounced in poetry, where meaning could be shaped by meter, word order, imagery, and cultural context, in addition to the semantics of the text. These challenges have motivated prior NLP work on poetry analysis (Kumar and Minz, 2014; Deshmukh et al., 2019), generation (Manurung, 2004; Agirrezabal et al., 2013), and translation (Ghazvininejad et al., 2017; Chakrabarty et al., 2021). Existing work on poetry translation generally considers the task of translating a poem from a source language into a poem in a target language. In contrast, we define *poem-to-prose translation*, the task of converting a metrically constrained poem into prose form within the same language or into a new language's prose. Taking 13th-17th Century Telugu language poems written in a metrically constrained manner as our source data, we propose this task both as intra-lingual (TE-TE) and cross-lingual (TE-EN) poem-to-prose translation (See Figure 1 for an overview of our setup). This task requires the model to arrive at a specific interpretation of the poem and express it in prose form. The output is not constrained by poetic form, but by the need to preserve the meaning of the input poem. The vocabulary and syntactical differences between poetry and prose, coupled with the evolution of the language over centuries makes the task challenging even in the intra-lingual setup. The cross-lingual setup additionally forces models to handle culturally grounded concepts that may not have direct translations in target languages, while conveying the intended meaning. Thus, whether it is intra-lingual or cross-lingual poem-to-prose translation, we argue that it is distinct from other seemingly related language generation tasks such as word reordering, paraphrasing, summarization, and text simplification.

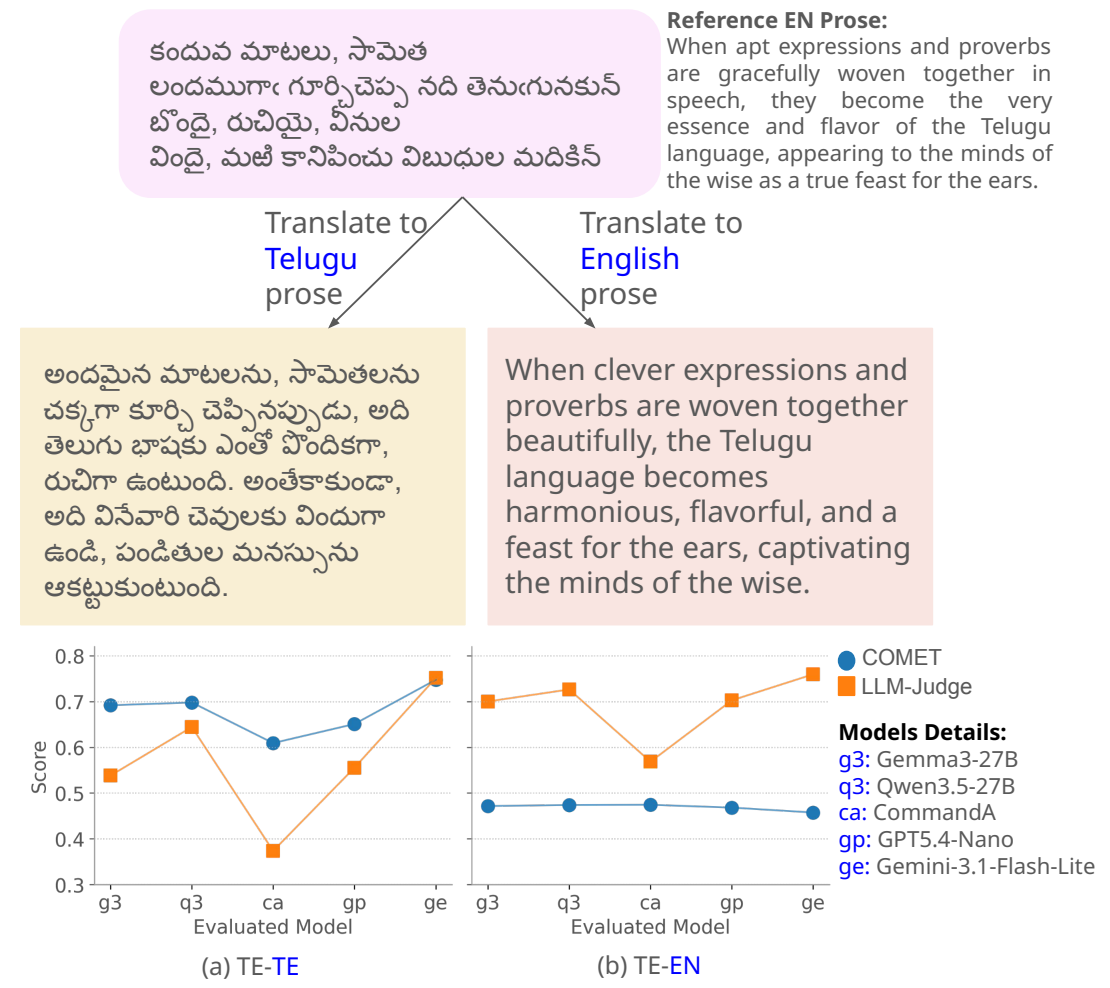

Figure 1: Overview of our task. Given a classical Telugu poem, different models generate Telugu and English prose translations. Outputs generated by Gemini-3.1-Flash-Lite model are seen here. The scatter plots compare reference-based COMET and normalized LLM-judge scores across models for TE–TE and TE–EN translation settings.

[1] https://huggingface.co/datasets/TeluguLLMResearch/Padyam2Gadyam

From a real-world standpoint, poem-to-prose translation is a common task while studying classical poetry in some Indian languages such as Telugu and Sanskrit, which have centuries-long literary history with the poetic form. It is also a part of the instructional material and evaluation tasks during high school education in some Indian school systems involving such languages. Thus, automated poetry-to-prose translation, while being a challenging task from a modeling perspective, could also help in making the study of classical poetry more accessible in such cases. With that as our motivation, we introduce పద్యం2గద్యం (Padyam2Gadyam), a benchmark dataset designed to assess the capabilities of today's state-of-the-art systems in understanding Telugu metrical poetry and generating prose translations for the same.

The dataset contains 600 manually curated poems from 13th to 17th century Telugu literary works along with their Telugu and English prose translations. Through this dataset, we study: (a) whether state-of-the-art language models can understand poems written in classical Telugu and generate meaning preserving prose translations in both Telugu and English, and (b) whether existing automatic MT evaluation methods can be used for this task. While the dataset focuses on Telugu and English, the task formulation can be extended to other languages and poetic forms in future.

**Contributions:** To summarize, our contributions are threefold. First, we introduce a new evaluation dataset for poem-to-prose translation in Telugu and English, which, to our knowledge, is the first such dataset in NLP research. Second, we benchmark two state-of-the-art machine translation systems and five proprietary and open-weight LLMs on this dataset. Third, we analyze the utility of automatic reference-free and reference-based metrics and LLM-as-a-judge evaluation for this task through both quantitative and qualitative analysis.

## 2 Literature Review

There has been a sustained interest in analyzing poetry in NLP research, focusing on a range of tasks such as rhyme detection (Hopkins and Kiela, 2017), meter identification (Lau et al., 2018), style analysis (Kaplan and Blei, 2007) and poem form analysis and visualization (Meneses and Furuta, 2025; Walsh et al., 2024). Poetry generation is also well-studied (Qu et al., 2026; Koziev and Fenogenova, 2025) including poetry generation from images (Xu et al., 2018). Within this space, the tasks closest to our interest are: poem summarization, poem translation and poem-to-prose reordering and paraphrasing.

**Poem Summarization:** Poem interpretation was studied as a summarization task for English free-verse, considering commentaries about those poems as summaries (Mahbub et al., 2023; Yousaf et al., 2025; Abir et al., 2026). In our case, the source poems are not contemporary free-verse but come from metrically constrained Telugu classical poetry from the 13th to the 17th century, where both the vocabulary and syntax structure are very different from contemporary Telugu prose. Further, we focus on prose generation both in Telugu and English.

**Poem Translation:** There is some research that focuses on taking a poem from a source language and translating/generating a poem in the target language (Chakrabarty et al., 2021; Wang et al., 2024; Resende and Hadley, 2024; Chen et al., 2025; Jamil et al., 2026). While these are related to our work in terms of the translation focus, they mainly focus on preserving poetic form in the target language, while we focus on generating prose translations that preserve the meaning.

**Poetry-to-Prose Reordering or Paraphrasing:** Another related task is poetry-to-prose reordering, where the output is gen-erated by reordering the words in the poetic input (Krishna et al., 2019; Das et al., 2025). Note that in this case of reordering, the the output differs from the input only in word order, not in vocabulary, unlike our problem definition where the output differs from input both in vocabulary and syntax, and the output is also not restricted to the same language.

To our knowledge, the closest related work to the current approach is the recent paper by Kalhor and Yaghoobzadeh (2026), who propose generating prose variations for input Persian ghazal couplets. While this work is similar to our setting, as the authors specify, the classical and contemporary forms of the language in their paper are still mutually intelligible, and this task is closer to paraphrasing with substantial vocabulary overlap between source and target. Further, while they focus on generating prose in a single language, we focus on generating both intra-lingual as well as cross-lingual prose translations. Finally, while their work uses 100 poems in the same Ghazal meter by a single poet as their source data, our dataset is larger and more diverse in terms of sources and authors, in addition to introducing a new, previously under-studied language into this space.

## 3 The పద్యం2గద్యం Padyam2Gadyam Dataset

To facilitate poem-to-prose translation in both Telugu and English, we construct the పద్యం2గద్యం dataset. The dataset consists of 600 samples, where each sample contains a Telugu poem paired with its Telugu and English prose translations. In addition to the prose translations, each sample includes information about the poem's meter and the source URL as shown in Figure 2.

**Dataset Construction:** Telugu belongs to the Dravidian family of languages, and is one of the recognized classical languages of India (Press Information Bureau, 2024) with over 1000 years of recorded literary history. Most of the ancient and medieval writings of Telugu are written in the form of metrically constrained poetry, which form the source data for our dataset.The poems were sourced from publicly available online archives, including Archive.org [3] and Telugu Wikisource [4], from books that have the poems along with its prose form in Telugu. We include a poem in the dataset only when both the poem and its Telugu prose version are available from the same source. To include poems from different time periods, we shortlist poems from the 13th to the 17th century from the available resources which have both poem and prose versions. The poems are of varying degrees of complexity, with the 13th century Sumathi Satakam supposed to be written in the lay person language (of that time) to the 16th century Manu Charitra written for more scholarly consumption, with heavy Sanskrit influence. Table 1 presents some statistics about the source data. All poems appear as a part of longer poetic compositions, and in terms of topics, the poems chosen for this dataset primarily consist of descriptions of nature, brief conversations between characters, verses on ethics and conduct, and descriptions of gods and goddesses.

The authors of this paper, who are native Telugu speakers, manually verified the poems and Telugu prose summaries for spelling, punctuation, OCR errors etc, and checked that no

**Poem:**
కనకపు సింహాసనమున
శునకము గూర్చుండబెట్టి శుభ లగ్నమునం
దొనరగ బట్టము గట్టిన
వెనుకటి గుణమేల మాను వినరా సుమతీ

**TE translation:**
కుక్కను తీసుకొని వచ్చి
మంచి ముహూర్తమునందు
బంగారు గద్దె మీద
కూర్చుండబెట్టి
పట్టాభిషేకము చేసినప్పటికి
దాని సైజగుణము నేలాగున
మానలేదో, ఆ విధముగనే
అల్పుడైనవానికి ఎంత
గౌరవము చేసి మంచి
పదవొసంగినను తన
నీచత్వమును వదలనేరడు.

**EN translation:**
Even if a dog is seated on a golden throne at an auspicious time, and crowned, its inherent virtues are not destroyed. In the same way, no matter how much respect a lowly person is shown and given a good position, he cannot abandon his pettiness.

**Metadata:**
Poem Meter (కందం [EN: Kandam]), Source (Sumati Satakam) and URL
https://te.wikisource.org/wiki/%E0%B0%B8%E0%B1%81%E0%B0%AE%E0%B0%A4%E0%B1%80_%E0%B0%B6%E0%B0%A4%E0%B0%95%E0%B0%AE%E0%B1%81

Figure 2: Dataset Example

[3] https://archive.org/
[4] https://te.wikisource.org/wiki

| Source Text | Time Period | # Samples | avg. # tokens | [min-max] # tokens |
|---|---|---|---|---|
| Andhra Bhagavatham | 15th Century | 250 | 77.81 | 46–196 |
| Molla Ramayanam | 16th Century | 100 | 75.61 | 45–118 |
| Manu Charitra | 16th Century | 74 | 80.59 | 47–113 |
| Sumati Satakam | 13th Century | 56 | 51.75 | 42–63 |
| Kalahasteeswara Satakam | 16th Century | 49 | 92.73 | 81–108 |
| Bhaskara Satakam | 16th Century | 44 | 92.55 | 84–104 |
| Dasarathi Satakam | 17th Century | 27 | 90.68 | 75–114 |
| | Total | 600 | | |

Table 1: Test Dataset Statistics
(Note: token counts come from XLM-R tokenizer [2]).

poem needs additional context beyond its own text for generating a prose translation. The language used in the prose is substantially different from the language used in the poetic form, and the surface character level overlap between Telugu prose and poetic forms in the dataset was only 26.16 as per ChrF (Popović, 2015), the standard character n-gram overlap measure used in machine translation research to compare two strings in the same language i.e., a reference translation and a machine translated output. Thus, although both the poem and prose versions contain the same information in terms of the overall meaning, they have a low ChrF score, confirming the substantial amount of stylistic and surface variation in their form.

Since there were no existing gold standard English prose translations for these poems and building them from scratch is out of reach considering the substantial human expertise it would require, we followed a semi-automatic approach to build them. We first used Google Translate to generate English prose translations from the gold-standard Telugu prose translations, then manually post-edited them to form our final gold standard English prose translations [5]. The post-editing process involved two annotators and focused on correcting grammatical errors and meaning errors, if any, and did not focus on changing the stylistic aspects, as we could expect substantial stylistic variation in the system generated outputs, given that none of the existing models are explicitly trained with this task.

While most of the machine generated translations are of good quality, there were some instances with incorrect translations, incomplete sentences and truncated text, which were corrected during the manual post-editing phase (see Table 4 in Appendix A for some examples). The average ChrF between the pre- and post-edited versions was 80.45, which is indicative that there was some degree of editing of the machine translated output. Note that while we used the Telugu prose to get the machine generated English prose translation and do the post-editing, the actual evaluation setup has only the original poem as input for both Telugu as well as English prose translation. The Telugu prose was only used as an anchor to build the gold standard English prose dataset.

Since we sourced the data from online archives, the possibility of data contamination among the evaluated LLMs cannot be ruled out, and it is a common issue running across all research that evaluates contemporary LLMs. However, it is important to note that the Telugu-English prose translation pairs were created as part of this work and did not exist in the original sources. Overall, the resulting పద్యం2గద్యం benchmark dataset[6] comprises of 600 Telugu poems with 600 Telugu prose and 600 English prose translations. It spans seven data sources spanning five centuries and covers diverse poetic meters. While we did not use the source and meter information in our current analysis, the dataset offers a potential to study how the variation in source, period, meter, and poetic style can affect the translation quality in future.

[5]We used the GOOGLETRANSLATE(text, [source_language], [target_language]) function in Google Sheets for the first part.

[6]https://huggingface.co/datasets/TeluguLLMResearch/Padyam2Gadyam

## 4 Experimental Setup

We formulate our task as poem-to-prose translation, where, given a Telugu poem as input, the model is required to generate a prose translation that preserves the meaning of the poem in two target language settings –TE and EN. The input poem remains the same in both settings, and only the target prose language changes.

**Model Selection** We evaluate two commercial machine translation systems and five LLMs on the పద్యం2గద్యం (PADYAM2GADYAM) dataset for the poetry-to-prose translation task. The machine translation systems are Google Translate [7] and Sarvam Translate [8], with the latter being a a state-of-the-art model specifically for translating to and from Indian languages. The five LLMs include two proprietary models, GPT-5.4-Nano [9] and Gemini-3.1-Flash-Lite [10], and three open-weight models of varying sizes, Gemma-3-27B (Team et al., 2025), Qwen-3.5-27B (Team, 2026), and CommandA-111B (Aakanksha et al., 2025). Among these models, Qwen-3.5 and CommandA explicitly mention support for Telugu, while Gemma3 reports evaluations using datasets covering Telugu. For the remaining two models, explicit support for Telugu is not specified. Note that the machine translation models do not generate translations for TE-TE scenario, as they expect languages with different ISO codes as source and target. So we report results for those models only in the case of TE-EN translation. All models are evaluated in a zero-shot setting to generate either Telugu or English prose translation of the input poem (Information about the exact prompt can be seen in the Appendix B).

### 4.1 Evaluation Metrics

Our source text is a metrically constrained Telugu poem, and the target text is a prose translation of the poem into Telugu and English. The word order between poetry and prose is very different, and while Telugu prose follows a Subject-Object-Verb order, English prose typically follows Subject-Verb-Object order. Further, while there is only one intended meaning for each poem, the reference translation is only one of the possible prose versions as many stylistic variations can exist. Therefore, machine translation metrics based on word, character, or ngram overlap, such as BLEU (Papineni et al., 2002), ChrF (Popović, 2015) as well as the lexical overlap focused summarization metrics such as ROUGE (Lin, 2004) are not suited for evaluating this task. Other recent research on Persian Ghazal to prose generation (Kalhor and Yaghoobzadeh, 2026) also showed that these metrics are unsuitable for comparing between poetry and prose forms.

Hence, we used COMET based metrics to evaluate the quality of the model generated outputs, as we formulated the task as a translation task. The COMET metric is primarily intended to be used for cross-lingual machine translation, where the source and target languages are different from each other, which is not the case of the TE-TE poem-to-prose task. However, considering that it aims to capture semantic similarity in a multilingual feature space, we can consider it as an appropriate metric even for the intra-lingual translation scenario. We used COMET-22 (Rei et al., 2022a) [11] for reference-based evaluation, where the source poem, model output, and reference prose translation are considered for evaluation, and COMETKIWI (Rei et al., 2022b) [12] for reference-free evaluation, where the source poem and model output are considered for evaluation without the reference translation. Telugu is listed as one of the languages supported by both the COMET models.

In addition to COMET-based evaluation, we take inspiration from recent MT research (and NLP research in general) in using LLM as a Judge (Kocmi and Federmann, 2023) and use Anthropic's Claude Haiku 4.5 [13] as the judge model. The judge evaluates both Telugu and English prose translations with respect to the input poem and assigns a score on a scale of 0-100 (the higher the better), along with explaining its reasoning. Similar to the COMET based evaluation, the LLM judge is used in both reference based and reference free settings. In the reference based setting, the judge is given the input poem, the model output, and the reference translation. In the reference free setting, the judge is given the input poem and the model output. The choice of the judge model is motivated by the preference to use a model outside the families of LLMs used to generate the translations. The judge prompts are available in Appendix B.

[7] https://translate.google.com/
[8] https://www.sarvam.ai/blogs/blog-mayura
[9] https://openai.com/index/introducing-gpt-5-4/
[10] https://deepmind.google/models/model-cards/gemini-3-1-flash-lite/
[11] https://huggingface.co/Unbabel/wmt22-comet-da
[12] https://huggingface.co/Unbabel/wmt22-cometkiwi-da
[13] https://www.anthropic.com/claude/haiku

Finally, we also do qualitative analysis on a subset of the data to compare among the models, check the correctness of the model outputs, the accuracy of the evaluation metrics, and the judge reasoning, and summarize some observations on the limitations of the existing evaluation setup for this task.

Both generation settings use prompts written in English and the task-specific prompts for each setting are available in Table 5 in Appendix B. All LLMs (generator models and the judge model) are accessed through OpenRouter [14] API through ReAct calls with DSPy [15]. The total API cost for these experiments is approximately 50 USD.

## 5 Results

*Can LLMs understand poems written in classic poetic Telugu?* We explore this question by comparing model performance in two experimental setups: a) TE-TE poem-to-prose translation, which assesses the intralingual poem-to-prose translation abilities of the models and b) TE-EN poem-to-prose translation, which explores the cross-lingual abilities of the models for this task. Table 2 and Table 3 present evaluation results for TE-TE and TE-EN cases respectively. For both settings, we report scores under reference-free and reference-based evaluation, using Comet scores and LLM-as-a-Judge (with Claude Haiku 4.5 as the judge model).

**TE-TE Translation:** As Table 2 shows, Gemini-3.1-Flash-Lite achieves higher score among the evaluated models in reference based evaluation using either Comet score or the LLM judge, with the LLM judge scores for the model being nearly identical in both reference-based and reference-free setups. In the reference-free setup, we observe that the Comet scores for all models are very close to each other, ranging from 0.45–0.47. While previous research on Comet scores shows that even small differences in scores can mean differences in human preferences (Kocmi et al., 2024), it is also possible in this case that the metric is unable to distinguish between model outputs in the absence of a reference, either because its training data does not include classical Telugu, or because of the intra-lingual setup. However, it is important to note that the Comet scores are generally higher and more varied in reference-based setting compared to the reference-free setting (despite the intra-lingual setup, that is).

**TE-EN Translation:** For TE-EN poem-to-prose translation, we observe that the two MT models perform poorly compared to the LLMs according to all metrics except the reference free Comet. This is expected, as such models are typically trained on shorter, sentence-level prose data and poetry is probably not a part of the training mix. The performance of LLMs follow similar trends as the TE-TE case, as Table 3 shows. Gemini-3.1-Flash-Lite achieves higher score in the reference-based Comet evaluation and in both reference-based and reference-free LLM-Judge evaluation. In the reference-free Comet setting, Qwen-3.5 achieves higher score, closely followed by Gemini-3.1-Flash-Lite. Similar to the TE-TE setting, the LLM-Judge scores are higher in the reference-free setting compared to reference-based setting, and they are on an average lower than the TE-TE, in both settings. A potential reason could be because of the presence of cultural references with no clear English equivalents, which we briefly explore in qualitative analysis.

In most cases, the LLM generated translations were longer than the reference translations, with GPT-5.4-Nano generating the longest translations and Qwen3.5-27B generating shortest translation on an average (See Table 7 in Appendix for the exact numbers). We did not analyze further on the relation between the data source and translation length, but focused primarily on the evaluation of transla-tion quality in this paper.

[14] https://openrouter.ai/. See Table 6 in Appendix for model urls.
[15] https://dspy.ai/

| | Reference-Based | | Reference-Free | |
|---|---|---|---|---|
| | COMET-DA | HAIKU-JUDGE | COMET-KIWI | HAIKU-JUDGE |
| Gemma3-27B | 0.6917 | 53.81 | 0.4716 | 70.00 |
| Qwen3.5-27B | 0.6978 | 64.42 | 0.4732 | 72.63 |
| CommandA-111B | 0.6091 | 37.35 | **0.4743** | 56.84 |
| GPT-5.4-Nano | 0.6508 | 55.53 | 0.4680 | 70.25 |
| Gemini-3.1-Flash-Lite | **0.7474** | **75.13** | 0.4574 | **75.94** |

Table 2: Results for TE-TE Poem-to-Prose Translation

| | Reference-Based | | Reference-Free | |
|---|---|---|---|---|
| | COMET-DA | HAIKU-JUDGE | COMET-KIWI | HAIKU-JUDGE |
| Google Translate | 0.5284 | 16.21 | 0.4701 | 31.33 |
| Sarvam Translate | 0.5573 | 28.66 | 0.4951 | 39.42 |
| Gemma3-27B | 0.6389 | 45.42 | 0.4828 | 65.65 |
| Qwen3.5-27B | 0.6613 | 59.83 | **0.5120** | 68.96 |
| CommandA-111B | 0.6036 | 32.91 | 0.4905 | 59.78 |
| GPT-5.4-Nano | 0.5977 | 41.76 | 0.4413 | 63.72 |
| Gemini-3.1-Flash-Lite | **0.7007** | **74.49** | 0.5070 | **72.51** |

Table 3: Results for TE-EN Poem-to-Prose Translation

The quantitative results raise a few questions about the automatic evaluation metrics. The COMET reference-free scores are relatively close to each other across models in both language settings, compared to the reference-based evaluation, which could indicate that the metric may not separate model-generated text well for this task without a reference translation when the input text is Telugu poems. In contrast, the LLM-Judge scores are able to show a wider score range in both reference based and reference free settings. The relatively large drop in the average LLM judge scores between both settings for all models except Gemini-3.1-Flash-Lite could be due to the gaps in the judge model's own understanding of Telugu poems which could have made it difficult to score model generated outputs with some meaning errors in them. The overall ranking of the models appears largely consistent across all cases, except when reference-free COMET scores were used to compare, though.

These results lead us to a conclusion that LLM judge based evaluation metrics may be more suitable for this task in a reference-free setting, while either COMET or LLM judges can be used when there are reference translations available, in order to understand the overall ranking of different models. LLM judge scores appear to offer a better way to infer the translation quality of individual models compared to COMET, as they are able to provide a wider score range.

### 5.1 Qualitative Analysis of LLM Generated Translations

To understand the quality of translations from a human reader perspective and understand the metrics better, we took a sample of 25 poems and their TE and EN prose translations from all LLMs (25*5 prose translations in TE, 25*5 translations in EN for each poem) and two human annotators (the authors, who are fluent native speakers of Telugu) rated the best model generated translation for each poem. We excluded the MT systems from this analysis as the translations were largely unreadable. Although we initially started with the aim of ranking the model outputs, we concluded that identifying the best model generated translation(s) and comparing with the automated metrics is a more realistic way of performing this analysis, instead of ranking all translations for each example. So, for each input poem, each annotator identified the model with best translation, and we compared the

**Input Poem**

ఎడపక దుర్జనుండొరుల కెంతయుఁ గీడొనరించుఁ గాని యే
యెడలను మేలు సేయఁడొక యించుకయైనను; జీడపుర్వు దాఁ
జెడఁ దిను నెంతెకాక పుడుసెండు జలంబిడి పెంప నేర్చునే
పొడవగుచున్న పుష్పఫల భూరుహ మొక్కటినైన **భాస్కరా** !

**EN Prose (Reference)**

Just as a worm eats away a well-grown fruit tree, but cannot water a tree enough to make it grow, so a wicked person learns to sow harm for others, but does no good to anyone.

**Model Outputs**

**Gemini-3.1-Flash-Lite**

**Oh Bhaskara!** A wicked person incessantly causes great harm to others, but never does even the slightest bit of good at any time. Just as a pest only consumes and destroys the plant it lives upon, can it ever learn to nurture a tree to grow tall and bear flowers and fruits?

**Qwen3.5-27B**

**O Sun!** A wicked person constantly inflicts harm upon others to the utmost extent, yet never performs even a single act of kindness in any way. Just as a thorny plant only breaks and consumes, does it know how to nurture a new sprout with water so that even one flower and fruit-bearing tree may grow tall?

**Gemma3-27B**

Even if one were to constantly praise a wicked person, they would not perform even a single act of kindness or benefit to anyone. Unlike them, the sun, **O Sun God,** tirelessly provides sustenance – it consumes the withered and decaying, and nurtures growth, causing a single plant to flourish with a multitude of flowers and fruits. It continuously bestows its light and warmth, fostering life and abundance. The wicked, however, are devoid of such generosity, offering no benefit whatsoever.

**CommandA-111B**

Even if the wicked may try to harm, they cannot diminish the glory of the virtuous. The mighty one, who rises high in the sky, remains unshaken. From ancient times, the earth has been adorned with fresh, green leaves, sparkling water, and lush growth, nurturing the blooming flowers and fruits. **O Bhaskar (the sun god),** you are the embodiment of this vibrant and fertile world, shining brightly and sustaining all life.

**GPT-5.4-Nano**

**O Bright Lord Bhaskara!** Even if there are people who behave like hardened villains and do not understand what is right, the good that needs to be done does not fail to come about. If one simply delays without learning and setting the proper effort, it will be hard for anything beneficial to be attained; yet by taking up the work rightly, improvement is possible. As the sun shines, growth follows—flowers and fruits, and even the shoots on creepers, sprout and flourish; therefore, your light and influence deserve praise, O Bhaskara!

Figure 3: Model translations with reference prose. Bold text shows how models translate the poem's refrain; underlined text highlights hallucinated content not present in the input poem.

agreement between the annotators. The annotation process is discussed in detail in Appendix F.

While there are cases where multiple translations were considered "good" by the annotators (without a specific ranking), there is a 95% agreement for TE-EN, and 100% for TE-TE for the best model generated translation (Gemini-3.1-Flash-Lite). The fact that the same model was the best scoring model across 3 of the 4 used metrics in Tables 2 and 3 indicates that there is a general agreement between humans and metrics on what the best model for this task is. Interestingly, for 3 poems in the sample, both annotators noted that none of the models have the perfect translation in both TE-TE and TE-EN cases. In these cases, all models failed to translate one phrase correctly in each poem, resulting in some missing meaning, although the text read fluently. All the metrics gave high scores for these translations, though.

Across models, we observe errors such as literal translation of poetic expressions, hallucinated content, omission of key meaning, and incorrect interpretation of culturally grounded terms, as shown in Figure 3 [16], which was not always captured by the metrics.

**Limitations of the Evaluation Metrics:** Both COMET and LLM judge, especially in the reference-free setup, are potentially biased towards scoring model hypotheses that use similar vocabulary to the one in source higher. There were examples of cases where the metrics assigned a high score to translations that superficially match the vocabulary of the source, but substantially deviate from the original meaning when taken as a whole. In all such cases, a human judge would assign a 0 score. This is not as prominent in the reference-based setup, for TE-EN case, and seemed to exist more in the TE-TE case even in the reference-based LLM judge setup, which was the most consistent metric among the four we used (see Table 13 in the Appendix for an example). The language of older Telugu texts may not have been a part of COMET training, which can explain this issue for those metrics. For the LLM judge, more detailed prompts, with few-shot examples, can be explored. Comparisons with other judge models would also help in developing a more comprehensive metric for this task. We leave that as a potential future research direction.

[16] Table 8–10 in Appendix show some examples of translations generated across various LLMs for three poems along with evaluation scores.

**LLM Judge's Reasoning Trace May be Useful:** Since we prompted the LLM judge model to "reason and act", the judge model's reasoning may offer some insights into how the model is evaluating. In one case, the reference-free LLM judge evaluation assigned the model generated translation a high score, but the judge reasoning had comments such as: *"The connection between different clauses in the original .... is not entirely clear in the translation"*. Despite such observations, it eventually scored the model as: "the meaning is substantially preserved..". For the same sample, in a reference-based evaluation, the same judge assigns a low score, rightly identifies that the machine generated output "*presents a significantly different meaning*" and highlights several other elements of hallucinations and interpretations in the translation [17]. Future research can focus on pattern-based analysis of judge reasoning to identify ambiguities or comprehension difficulties, which may also help gauge the confidence of an LLM judge's decision.

## 6 Summary and Discussion

We proposed a poem-to-prose translation task that formulates prose generation as translation of a poem into prose form, and introduced పద్యం2గద్యం(PADYAM2GADYAM), a benchmark dataset containing Telugu and English translations of 600 Telugu poems from 13th to 17th century texts. Evaluations using this dataset with two MT systems and 5 LLMs indicate that while the MT systems perform poorly, some LLMs are capable of generating, fluent, and correct prose translations (into Telugu and English), and even medium sized models (27B) exhibit some knowledge of the task. Although the automated metrics are consistent about the overall ranking across models, they seem to capture high level semantic similarity, and fail to identify hallucinations and missing information in the generated outputs.

Among the metrics, LLM judge appears closer to human preference in the reference based evaluation setting. However, in its current form, it is not completely free from the above mentioned problems either. Calibrating the LLM judge by making use of the reasoning trace patterns, and by giving few-shot examples with more detailed prompts, is a possible direction to explore in future. Focusing on increasing the dataset size to be sufficient enough for few-shot prompting or supervised fine-tuning is also another direction to consider. Note that several resources exist with prose translations of classical Telugu literature, and a larger scale effort to build such datasets is not impossible, subject to availability of human resources. The results of this study should be considered as a starting point in making classical Telugu poetry more accessible to contemporary audience, and not as an end in itself. Hopefully, this would prompt similar research in other languages, enabling comparative studies in future.

## Limitations

The relatively small sized data sample and restricting to a zero-shot evaluation setup, the choice of models used, the usage of a single LLM judge, and the process of creating the English prose translations can all be identified as potential limitations of this study, which we discuss briefly below:

1. **Dataset size**: While the dataset size is small to do any task specific fine-tuning right now, it has to be noted that it still provides a reasonably sized sample to study the capabilities of current models in a relatively new generation task that re-quires reasoning about the poetic form in a low resource language. Considering that there is a large amount of Telugu classical studies literature available online, it is not impossible to develop a larger dataset in future, subject to the availability of human and financial resources.

[17]Full reasoning trace is in Appendix Table 11 and another example with Gemini-3.1-Flash-Lite is in Table 12.

2. **Limited set of models and a single prompt setting:** Some obvious limitations to the experiments described in this paper are the limited set of LLMs used, and the lack of further exploration of the prompt variations and other prompting strategies. While the choice of models and the number of models used for evaluation do not cover the whole space of available models, they still cover models of various sizes, and aim to achieve a balance between open-weight and proprietary models. The goal of the research is to develop the task framing and introduce a dataset relevant to the task, which was appropriately addressed in the paper. Considering that the dataset will be released publicly, with unlimited budget, it is straight forward to access other models and compare the results, if needed. Similarly, more models can be explored as LLM judges as well. We view these as potential extensions to broaden the coverage of models and prompt variations, but not as limitations of the task framing and dataset development process itself, which are the main contributions of this paper.

3. **English prose translations:** The use of post-editing based approach for the English data creation is not ideal, but it is important to note that it is a common real-world practice in the translation services industry as well, as constructing fully manually created translations (English prose versions for these poems in our case) from scratch will need substantial amount of human expertise and financial effort. We view this as a realistic starting point for resource creation, given the recent advances in machine translation. Although there may be a possibility that this process may induce a bias towards machine translated outputs, we found no evidence for that in our results so far, and the anchor model used to build the English dataset itself was the worst performing model in the TE-EN translation task (Table 3).

## Acknowledgements

We thank Rebecca Knowles and Gabriel Bernier-Colborne for their feedback on the initial draft, Jackie Lo for the guidance on metrics and evaluation, and the reviewers for their detailed comments. Kranti has been funded by the Potsdam Graduate School (POGS) and Completion Scholarship from Department of Linguistics, University of Potsdam.

# A Dataset Curation - Editing EN Translation

Table 4 shows some examples of post-editing of EN prose. Note that the human post-editors had access to both the original Telugu poem as well as the gold standard Telugu prose translation.

# B LLM Prompts for Translation and Evaluation

Table 5 shows the all the prompts used for generation and LLM Judge evaluations. Table 6 shows the details about the LLMs used for generation and evaluation.

# C Example Translations from LLMs

Table 7 shows some statistics on translation length. Figure 3 and Tables 8-9 contain example outputs from all the five LLMs for TE-TE and TE-EN translations. For TE-EN, two examples are shown, to illustrate one case where all LLMs get the overall meaning, and another one where most of them hallucinate. The source poem and its reference output in the target language are provided in both Tables.

| Google Translate Version | Human Post-Edited Version |
|---|---|
| I am asking you something with awe. You have seen the strange things that I have seen, but it has taken me years to come to see them. Your face tells me that you are very young. At such a young age, you have been able to see everything, aha! Your greatness is not only a source of pride for you, but also a source of pride for us! *((Translation matches the vocabulary, but misses the meaning)* | I am asking you something without hesitation. You have seen many strange things, but it will take me years to go and see them even if I have wings and if I can fly. Your face tells me that you are very young. At such a young age, how have you have been able to see everything! You know the secret of your greatness, but how can we know? |
| The birds flew from their nests to their nests, the sun was shining brightly, the stars were *(Incomplete Translation)* | The birds returned from their feeding grounds in long rows to their nests. The star ruby stones that were heated up under the sun's head started cooling down. Mirages started fading away. Light from all directions started dimming. The lotus flowers began closing their petals. The round sun started turning crimson. |
| The fire that burns in the fire is burning in the fire. Just because you drink milk, do you have good qualities? *(Translation completely missed the vocabulary, meaning and the cultural context.)* | The god of fire, whose birth place is the ocean, can burn the whole earth without mercy. Why should I think you will have good thoughts, O moon god, just because you were born in a sea of milk? *(Note: Fire's birth place being the ocean, and Moon's birth place being the sea of milk are culturally grounded references)* |

Table 4: Examples of Post-Editing the Machine Translated Output for Dataset Construction

| | |
|---|---|
| Translation Prompt | Translate the given poem in classical Telugu into [Telugu/English] prose. |
| Reference-Free LLM Judge | Score the given machine generated translation from a given Classical Telugu poem to *target_lang* prose on a continuous scale from 0 to 100 where a score of zero means "no meaning preserved" and score of one hundred means "perfect meaning and grammar". You will be given a classical Telugu poem and its translation into *target_lang* as inputs. Just give a single score in the range of 0-100. |
| Reference-Based LLM Judge | Score the given machine generated translation from a given Classical Telugu poem to *target_lang* prose on a continuous scale from 0 to 100. You will be given a source poem, a reference human translation, and a machine generated output. Note that the reference human translation is meant to serve mainly as a reference for meaning preservation, and the style of writing in the machine generated output may be different from the reference. A score of zero means "no meaning preserved" and score of one hundred means "perfect meaning and grammar". Just give a single score in the range of 0-100. |

Table 5: Translation and Evaluation Prompts

## D LLM Reasoning Traces

Tables 11 and 12 show judge reasoning for two EN translations generated by two LLMs, both in reference-based and reference-free settings.

## E LLM Judge Scoring Discrepancy

Table 13 shows an example where LLM judge gives high scores for poor translations.

## F Human Annotations in Qualitative Analysis

From the evaluation set, we randomly selected 25 samples. Each sample includes a poem, ground-truth TE and EN prose, along with responses generated by five models. Annotators are shown the poem and the corresponding ground truth, followed by the five model responses anonymized as Model1, Model2, Model3, Model4, and Model5. They are asked to select the response that best reflects the poem's intent by indicating the corresponding model number (e.g., 3, 2, or 5). If multiple responses are considered good, annotators rank them from best to next best (e.g., 3,4,1). If none of the responses are suitable, they select *None*. Overall, the annotators show 100% agreement for TE-TE and 95% agreement for TE-EN, with Gemini-3.1-Flash-Lite selected as generating the best response.

| Model | OpenRouter URL | Release Date |
|---|---|---|
| Gemma3-27B | https://openrouter.ai/google/gemma-3-27b-it | Mar 2025 |
| Qwen3.5-27B | https://openrouter.ai/qwen/qwen3.5-27b | Feb 2026 |
| CommandA | https://openrouter.ai/cohere/command-a | Mar 2025 |
| GPT-5.4-Nano | https://openrouter.ai/openai/gpt-5.4-nano | Mar 2026 |
| Gemini-3.1-Flash-Lite | https://openrouter.ai/google/gemini-3.1-flash-lite-preview | Mar 2026 |
| Claude-Haiku-4.5 | https://openrouter.ai/anthropic/claude-haiku-4.5 | October 2025 |

Table 6: OpenRouter Model URLs

| | TE-TE | | TE-EN | |
|---|---|---|---|---|
| **Data Source** | **avg #tokens** | **[min-max]** | **avg #tokens** | **[min-max]** |
| Ref. Translations | 84.28 | [30, 268] | 66.77 | [19, 242] |
| Gemma3-27B Generated | 105.67 | [37, 274] | 99.80 | [45, 256] |
| Qwen3.5-27B Generated | 77.52 | [3, 346] | 76.24 | [3, 233] |
| CommandA Generated | 97.13 | [40, 251] | 88.65 | [36, 264] |
| GPT-5.4-Nano Gen. | 131.12 | [50, 440] | 114.47 | [44, 303] |
| Gemini-3.1-Flash-Lite | 89.75 | [35, 248] | 80.55 | [37, 250] |

Table 7: Translation Length Statistics

| | |
|---|---|
| Original Poem | అనఘ! విను, రసజ్ఞులై వినువారికి<br>మాటమాట కధిక మధురమ్మైన<br>యట్టి కృష్ణు కథన మాకర్ణనము సేయఁ<br>దలఁపు గలదు, మాకుఁ దనివి లేదు. |
| Telugu Prose-Ref | పుణ్యాత్మా! సూతా! శ్రీకృష్ణుని కథలు ఆసక్తితో ఆలకించే రసజ్ఞుల హృదయాలకు పదే పదే మధురా-తి మధురాలై ఆనందాన్ని కలిగిస్తాయి. వాటిని ఆలకించాలని మాకు ఎంతో కుతూహలంగా ఉంది. ఎన్ని విన్నా మాకు తనివి తీరటం లేదు. |
| Gemma3-27B<br>[0.62, 0.75, 72, 62] | అనఘా! విను. రసజ్ఞులైన, కథలు వినడానికి ఆసక్తి గలవారికి, ప్రతి మాట మధురంగా ఉండే కృష్ణుని కథను మాకు వినిపించు. ఆ కథను చెప్పడానికి మాకు శక్తి లేదు, కానీ వినడానికి మాత్రం చాలా ఆసక్తిగా ఉన్నాము. |
| Qwen3.5-27B<br>[0.52, 0.68, 82, 72] | ఓ పాపం లేనివాడా! వినుము. రసజ్ఞులుగా వినువారికి, ప్రతి మాట అతి మధురంగా ఉండే అటు-వంటి కృష్ణుని కథనం మాకు వినిపించే ఉద్దేశం ఉంది. మాకు అందులో ఎప్పుడూ తనివి రాదు. |
| CommandA-111B<br>[0.54, 0.69, 62, 52] | అనఘ! రసజ్ఞులైన వారికి కృష్ణుని కథనాలు మాటలు మాటలు అత్యంత మధురమ్మైనవి. ఆ కథనా-లు మన చెవులకు సంతోషంగా వస్తాయి. అవి మనకు దలచేయడం లేదు, మనకు ఆ కథనాలను అనుభవించే సందర్భం లేదు. |
| GPT-5.4-Nano<br>[0.58, 0.68, 82, 74] | అనఘా! వినండి. రసజ్ఞులు వినే వారికి మాటమాటకీ కధికమ్మైన (ఆప్యాయంగా, మధురంగా ఉండే) అట్టి కృష్ణుని కథనాన్ని మేము వినడం (శ్రవించడం) తమ ప్రార్థన. దానికి సంబంధించిన ఆకాంక్ష మా మనసులో ఉంది; అయితే మాకు దినివి (అటువంటి అవకాశం/వనరులు) లేవు. |
| **Gemini-3.1-Flash-Lite**<br>[0.52, 0.81, 78, 82] | ఓ పవిత్రుడా! వినుము. రసజ్ఞులై (కథలోని మాధుర్యాన్ని ఆస్వాదించే శక్తి కలవారై) వినేవారికి, ప్రతి మాట కూడా ఎంతో మధురంగా ఉండే అటువంటి శ్రీకృష్ణుని కథలను వినాలని మాకు ఎంతో ఆసక్తిగా ఉంది. ఆ కథలను వింటూ ఉంటే మాకు ఎంత విన్నా తనివి తీరడం లేదు. |

Table 8: Example for TE-TE Translation. Scores under each model follow this order - [Comet-RefFree, Comet-RefBased, LLMJudge-RefFree, LLMJudge-RefBased]

| | |
|---|---|
| Original Poem | మొగము బిగించి, పాదములు మొత్తముగా వడినూఁది త్రొక్కి, నీ<br>టుగ మొగమెత్తి, భీకర కఠోర రవంబున నార్చి, బాహు ల<br>త్యగణిత లీల నూఁచి, వలయంబుగ వాలముఁ ద్రిప్పి, వ్రేఁగునన్<br>నగము సగమ్ము గ్రుంగఁ, గపి నాథుఁడు నింగికి దాఁటె జివ్వునన్. |
| English Prose-Ref | With his head firmly planted, he stomped his feet on the mountain, raised his face gracefully, gave a terrifying scream, swinged his shoulders dramatically, and twirled his tail in a circle, causing the mountain to collapse halfway under his weight, Hanuman soared into the sky with great speed. |
| Gemma3-27B<br>[0.45, 0.71, 72, 76] | With a determined face and feet firmly planted, Hanuman pressed down and launched himself upwards. He raised his face towards his destination and emitted a terrifying, harsh roar. Displaying his immeasurable strength through playful and graceful movements of his arms, he circled and whirled like a whirlwind. Then, with great speed, he leaped across the sky, crossing over to Lanka in an instant. |
| Qwen3.5-27B<br>[0.52, 0.75, 78, 68] | Frowning fiercely, he pressed his feet firmly into the ground from heel to toe. Raising his head straight up, he let out a terrifying and harsh roar. Swinging his arms with uncountable grace and whirling his tail in a circle, the force of his movement caused the mountain to sink halfway into the earth. In an instant, the Lord of the Monkeys leaped into the sky. |
| CommandA-111B<br>[0.53, 0.68, 52, 35] | Lifting his face, he walked with heavy steps, his feet pounding the ground. Roaring with a fierce and harsh voice, he performed a dance-like movement, his arms waving in a carefree manner, and his garland swaying. With a resounding sound, he struck the ground with his foot, and the lord of the earth bowed down in submission. |
| GPT-5.4-Nano<br>[0.45, 0.66, 42, 58] | With his face tightened and his feet fully braced, he moves onward; he raises his mouth and sets loose a fearsome, harsh roaring sound. He performs the wondrous play of his arms with a spirit of abandonment, and as the encircling band swings and slips, the daring deed is accomplished. Thus the city and its loud crowd are made to surge and sway—so that the leader, the Nāthu, will rise and fly straight into the sky, as if pierced through the heavens. |
| Gemini-3.1-Flash-Lite<br>[0.49, 0.74, 72, 78] | Setting his face with determination and pressing his feet firmly against the ground, the lord of monkeys lifted his head with pride. He roared with a terrifying and harsh sound, swung his arms with boundless grace, and lashed his tail in a circular motion. With such immense weight that half the mountain sank beneath him, he leaped swiftly into the sky. |

Table 9: Example for TE-EN Translation where all LLMs generally understand the poem, although the specific details differ among the translations. Scores under each model follow this order - [Comet-RefFree, Comet-RefBased, LLMJudge-RefFree, LLMJudge-RefBased]

| | |
|---|---|
| Original Poem | ఎడపక దుర్జనుండొరుల కెంతయుఁ గీడొనరించుఁ గాని యే<br>యెడలను మేలు సేయఁడొక యించుకయైనను; జీడపుర్వు దాఁ<br>జెడఁ దిను నెంతెకాక పుడుసెండు జలంబిడి పెంప నేర్చునే<br>పాడవగుచున్న పుష్పఫల భూరుహ మొక్కటినైన భాస్కరా! |
| English Prose-Ref | Just as a worm eats away a well-grown fruit tree, but cannot water a tree enough to make it grow, so a wicked person learns to sow harm for others, but does no good to anyone. |
| Gemma3-27B [0.35, 0.59, 68, 52] | Even if one were to constantly praise a wicked person, they would not perform even a single act of kindness or benefit to anyone. Unlike them, the sun, O Sun God, tirelessly provides sustenance – it consumes the withered and decaying, and nurtures growth, causing a single plant to flourish with a multitude of flowers and fruits. It continuously bestows its light and warmth, fostering life and abundance. The wicked, however, are devoid of such generosity, offering no benefit whatsoever. |
| Qwen3.5-27B [0.41, 0.61, 72, 68] | O Sun! A wicked person constantly inflicts harm upon others to the utmost extent, yet never performs even a single act of kindness in any way. Just as a thorny plant only breaks and consumes, does it know how to nurture a new sprout with water so that even one flower and fruit-bearing tree may grow tall? |
| CommandA-111B [0.41, 0.52, 68, 22] | Even if the wicked may try to harm, they cannot diminish the glory of the virtuous. The mighty one, who rises high in the sky, remains unshaken. From ancient times, the earth has been adorned with fresh, green leaves, sparkling water, and lush growth, nurturing the blooming flowers and fruits. O Bhaskar (the sun god), you are the embodiment of this vibrant and fertile world, shining brightly and sustaining all life. |
| GPT-5.4-Nano [0.39, 0.57, 72, 28] | O Bright Lord Bhaskara! Even if there are people who behave like hardened villains and do not understand what is right, the good that needs to be done does not fail to come about. If one simply delays without learning and setting the proper effort, it will be hard for anything beneficial to be attained; yet by taking up the work rightly, improvement is possible. As the sun shines, growth follows—flowers and fruits, and even the shoots on creepers, sprout and flourish; therefore, your light and influence deserve praise, O Bhaskara! |
| **Gemini-3.1-Flash-Lite** [0.46, 0.70, 72, 82] | Oh Bhaskara! A wicked person incessantly causes great harm to others, but never does even the slightest bit of good at any time. Just as a pest only consumes and destroys the plant it lives upon, can it ever learn to nurture a tree to grow tall and bear flowers and fruits? |

Table 10: Example for TE-EN Translation where there is a varying degree of mistranslation and hallucination across LLMs. Only the Gemini model seems to generate a translation that is faithful to the original. Scores under each model follow this order - [Comet-RefFree, Comet-RefBased, LLMJudge-RefFree, LLMJudge-RefBased]

| | |
|---|---|
| GPT-5.4-Nano | O Bright Lord Bhaskara! Even if there are people who behave like hardened villains and do not understand what is right, the good that needs to be done does not fail to come about. If one simply delays without learning and setting the proper effort, it will be hard for anything beneficial to be attained; yet by taking up the work rightly, improvement is possible. As the sun shines, growth follows—flowers and fruits, and even the shoots on creepers, sprout and flourish; therefore, your light and influence deserve praise, O Bhaskara! |
| Reference-Free LLM Judge Reasoning (Assigned score: 72) | The hypothesis attempts to translate a classical Telugu devotional poem into modern English prose. Let me analyze the key aspects:<br>**Strengths:** 1. The translation captures the overall devotional tone addressing Bhaskara (the Sun God) 2. It preserves the ethical message about virtue and righteousness despite obstacles 3. The metaphor of the sun causing growth and flowering is maintained 4. The structure follows a logical progression from the source material 5. Grammar and prose structure are competent and readable in English<br>**Weaknesses:** 1. The source text appears to be a complex classical Telugu verse with specific poetic structure that is difficult to fully render in prose 2. Some nuances of the original may be lost - the exact phrasing about "hardened villains" and their understanding is somewhat interpretive 3. The connection between different clauses in the original (particularly around "జీడపుర్వు దాఁ జెడఁ దిను") is not entirely clear in the translation 4. The phrase about "proper effort" and "learning" seems to add some interpretation rather than strict translation 5. Some technical theological or poetic elements specific to classical Telugu devotion may not be fully captured<br>**Overall Assessment:** The translation demonstrates a solid understanding of the source material and conveys the primary message effectively. However, given the complexity of classical Telugu poetry and the loss of poetic form in conversion to prose, along with some interpretive liberties, it's a good but not perfect translation. The meaning is substantially preserved with reasonable English grammar and readability. |
| Reference-based LLM Judge Reasoning (Assigned score: 28) | To evaluate this translation from Classical Telugu to English, I need to analyze how well the machine-generated output (hyp) preserves the meaning of the source text (src) compared to the reference translation (ref).<br>The reference translation presents a clear moral teaching: just as a worm damages a fruit tree but cannot nourish it to help it grow, a wicked person learns to harm others but does no good to anyone. This is a straightforward analogy about the nature of malevolent actions.<br>The machine-generated output, however, presents a significantly different meaning: - It addresses "Bright Lord Bhaskara" (the sun god) - It discusses people who behave like villains but suggests that "good that needs to be done does not fail to come about" - It emphasizes the possibility of improvement through proper effort - It praises Bhaskara's light and influence - The overall tone is devotional and optimistic<br>Comparing to the source text (which appears to be a classical Telugu poem with references to "ఎడపక" (worm), "గీడొనరించు" (to eat/damage), and "పుష్పఫల" (flowers and fruits)), the reference translation captures the cautionary/moral lesson about wicked people's inability to do good.<br>The machine-generated output, while poetic and well-written in English, appears to: 1. Misinterpret or significantly expand the meaning of the source 2. Introduce religious/devotional elements (Bhaskara worship) that may not be the primary focus 3. Present an optimistic message about improvement rather than a warning about malevolent nature 4. Diverge substantially from the concise moral lesson in the reference<br>The grammar and prose quality of the machine output is good, but the semantic preservation is poor. The translation seems to have lost the core meaning of the source text. |

Table 11: Example of LLM Judge Reasoning for GPT-5.4-Nano's output for the example from Figure 3 Reference-based and Reference-Free settings)

| | |
|---|---|
| Gemini-3.1-Flash-Lite | Oh Bhaskara! A wicked person incessantly causes great harm to others, but never does even the slightest bit of good at any time. Just as a pest only consumes and destroys the plant it lives upon, can it ever learn to nurture a tree to grow tall and bear flowers and fruits? |
| Reference-Free LLM Judge Reasoning (Assigned Score: 72) | The translation attempts to convey a classical Telugu poem addressing Bhaskara (likely the mathematician/astronomer). Let me analyze the fidelity:<br>Strengths: - The translator captures the main philosophical argument: a wicked person cannot do good, just as a pest cannot nurture a tree - The metaphor of the pest consuming the plant is present and relevant - The addressing of "Bhaskara" is preserved - The overall structure of the comparison/rhetorical question is maintained - The concluding image of trees bearing flowers and fruits is captured<br>Weaknesses: - The Telugu original is quite compressed and poetic with specific grammatical constructions that are difficult to fully render - Some nuances may be lost in the transition from classical Telugu to modern English prose - The exact semantic relationships and particles in the original (like the emphatic markers and conditional structures) are somewhat flattened - The original's compressed style gives it force that prose expansion diminishes slightly - Some specific linguistic features of classical Telugu devotional/philosophical poetry are not fully preserved<br>The translation is substantially accurate in conveying the core meaning and intent - the critique of a fundamentally wicked person's inability to do good, using the pest-plant metaphor as evidence. The English is grammatically sound and idiomatic. While there is some loss of poetic compression and certain classical nuances, the translation succeeds in preserving the essential meaning and philosophical argument. |
| Reference-Based LLM Judge Reasoning (Assigned Score: 82) | Let me analyze the machine-generated translation against the source text and reference translation.<br>The source is a Classical Telugu verse that appears to be making a moral comparison about wicked people - just as a worm/pest cannot nurture a tree despite consuming it, so too a wicked person cannot do good despite learning to cause harm.<br>Comparing the hypothesis to the reference: - Both capture the core meaning: a wicked person causes harm but does no good - Both use the pest/worm analogy comparing destructive behavior to inability to nurture growth - The reference is more concise and prose-like - The hypothesis is more elaborate, using rhetorical questions and poetic devices<br>Strengths of the hypothesis: 1. Preserves the central moral message accurately 2. Maintains the pest/plant analogy from the source 3. The rhetoric matches the poetic nature of the original 4. Adds the element of addressing "Bhaskara" (the poem's addressee) 5. Expands on the contrast between harm and good-doing 6. Grammar is correct and English is fluent<br>Weaknesses of the hypothesis: 1. Somewhat more verbose than necessary 2. The rhetorical question format, while poetic, may be slightly interpretive beyond the source 3. Minor: "pest" vs the original's more general concept might be slightly narrower<br>Overall assessment: The translation preserves the meaning very well. The core philosophical point about wicked people being unable to do good, paralleled with a pest's inability to nurture, is maintained. The style differs from the reference (more poetic/rhetorical), but the judge prompt explicitly notes that style differences are acceptable. The meaning preservation is strong, grammar is excellent, and the cultural element (addressing Bhaskara) is appropriate. |

Table 12: Example of LLM Judge Reasoning for Gemini-3.1-Flash-Lite's output for the example from Figure 3 Reference-based and Reference-Free settings)

**Poem:**
వెలఁది యిచ్చటిసంపెంగవిరులు గహన
దేవతలయిండ్ల నెత్తినదీపశిఖలు
గాక యుండినఁ బంచబంగాళ మగుచుఁ
బాఱునే మత్తమధుకరపటలతమము.

**Reference:** చెలీ! ఇచ్చటి సంపెంగపూలు వనదేవతలయిండ్లలోని దీపజ్వాల లనఁదగును. కావుననే కాఱుతుమ్మెద లనుచీకట్లు వానినిఁ జేరక దూరముగాఁ బాఱుచున్నవి.

**Gemma3-27B Translation:** అక్కడ తెల్లని మల్లె పువ్వులు, బంగారు రంగులో ఉండే చంపక పువ్వులు సమృద్ధిగా వికసిస్తున్నాయి. అవి దేవతల ఇళ్ళ పైభాగంలో వెలిగే దీపాల వలె ప్రకాశవంతంగా ఉన్నాయి. ఆ పువ్వుల నుండి వచ్చే సువాసన అన్ని దిక్కులకు వ్యాపించి, ఒక మత్తు వాతావరణాన్ని సృష్టిస్తోంది. తేనె కోసం వచ్చిన తేనెటీగలు పువ్వుల చుట్టూ తిరుగుతూ మత్తులో ఉన్నట్లుగా సందడి చేస్తున్నాయి.

**Judge scores:** Ref-Free: 75, Ref-Based: 82

**Comment:** Translation has all the vocabulary from the original, but also has some content not in the original, and leads to a meaning that is different from the original.

---

**Poem:**
దేవ గణములార! దిక్పాల వరులార!
వృక్ష జాతులార! పక్షులార!
కుటిల దానవుండు గొంపోవుచున్నాఁడు
కరుణతోడ నన్నుఁ గావరయ్య!

**Reference:** ఓ దేవతల్లారా! దిక్పాలకులారా! వృక్షాల్లారా! పక్షుల్లారా! ఒక మోసగాడయిన రాక్షసుడు నన్ను ఎత్తుకుపోతున్నాడు, దయతో నన్ను రక్షించండి.

**GPT-5.4-Nano Translation:** దేవగణములారా! దిక్పాలకులవరులారా! వృక్షజాతులారా! పక్షులారా! కుటిలమైన దానవుడు గొం-పుగా చుట్టుకుని ఉన్నాడు. కరుణతో నన్ను కాపాడుమని (రక్షింపుమని) వేడుకొనుచున్నాను; నీవు దయచేసి నన్ను గావరయ్యా.

**Judge scores:** Ref-Free: 72, Ref-Based: 85

**Comment:** The translation largely repeats the same words from the original, and hence preserves the meaning. There is some hallucination (గొంపుగా చుట్టుకుని ఉన్నాడు).

---

**Poem:**
శస్త్రముఁ దాల్చినం దగునె? సన్నుతి కెక్కిన భార్గవుం డనన్
శాస్త్రము గాదు, నా కెదిరి సంగర భూమిని నిల్చినంతనే
శాస్త్ర ముఖంబులన్ నృపులఁ జక్కుగఁ జేయుఁదు గాఁన నిప్పుడున్
శస్త్రము శాస్త్రముం గలవు సాహస వృత్తిని రమ్ము పోరఁగన్.

**Reference:** నేను ప్రసిద్ధి గాంచిన, భృగువంశ బ్రాహ్మణుడనే. అయితే ఆయుధంపట్టి పోరాడడానికి తగనా? ప్రస్తుతము శాస్త్రచర్చకాదు ముఖ్యం. నాకు ఎదురుగా యుద్ధరంగంలో నిలిచిన రాజులను ముక్కలుగా ఖండిస్తాను. నా దగ్గర ఇప్పుడు శస్త్రము శాస్త్రము రెండూ ఉన్నాయి. ధైర్యంచేసి నాతో యుద్ధానికిరా!

**GPT-5.4-Nano Translation:** "శస్త్రాన్ని చేతబట్టినట్లే అది దగునా? సన్నుతి (ప్రశంస) చెప్పి మార్గం చూపిన భర్గవుడు అని అనేవాళ్ల మాట సరియైనది కాదు. అది శాస్త్రము కాదు; నా ఎదురుగా సంగరభూమిలో (యుద్ధభూమిలో) నన్ను నిలిపినంతనే శాస్త్ర ముఖాలు ఉన్నట్లుగా కనిపించే పనులు రాజులకు కూడా జక్కుగా (సులభంగా, గెలుపుగా) జరగవు. అయితే, శస్త్రం శాస్త్రం' రెండూ కలిసినవారు మాత్రమే సాహసవృత్తిని పుట్టించి, ముందు పోరాడతారు."

**Judge scores:** Ref-Free: 82, Ref-Based: 82

**Comment:** The hypothesis contains all the vocabulary of the original poem and the reference translation, but is hard to follow, and does not convey the meaning of the original poem.

Table 13: TE-TE examples where LLM judge gives high scores for poor translations, in reference-based and reference-free settings